\documentclass[11pt]{article}

\usepackage[preprint]{acl}

\usepackage{times}
\usepackage{latexsym}

\usepackage[T1]{fontenc}

\usepackage[utf8]{inputenc}

\usepackage{microtype}

\usepackage{inconsolata}

\usepackage{graphicx}

\usepackage{textcomp}

\usepackage{booktabs}
\usepackage{xcolor}
\usepackage{amsmath}
\usepackage{amssymb}
\usepackage{tikz}
\usepackage{enumitem}
\usetikzlibrary{positioning,arrows.meta,shapes.geometric}
\usepackage{listings}
\usepackage{xspace}
\usepackage{marvosym}
 
\usepackage[most]{tcolorbox}
\newtcolorbox{promptbox}[1]{%
  colback=gray!7, colframe=gray!65!black, boxrule=0.4pt, arc=1mm,
  left=2mm, right=2mm, top=1mm, bottom=1mm, breakable,
  title={#1}, fonttitle=\bfseries\footnotesize, coltitle=black,
  colbacktitle=gray!20, before skip=6pt, after skip=6pt}
\newcommand{\schema}[1]{{\ttfamily\scriptsize #1}}
 
\newcommand{\td}[1]{\textcolor{red}{\textbf{#1}}}
\newcommand{\sys}{\textbf{CoLearn}\xspace}

\title{CoLearn: An Agentic Tutor that Learns its Learner\\in a Human--AI Co-Learning Loop}

\author{
 \textbf{Kailai He},
 \textbf{Zhihao Wu},
 \textbf{Linhai Zhang},
 \textbf{Runcong Zhao},
 \textbf{Yulan He},
 \textbf{Jiazheng Li}
\\
 King's College London
\\
 \small{
  \texttt{hekailai@gmail.com}
 },
 \small{
  \texttt{\{zhihao.wu, linhai.zhang, runcong.zhao, yulan.he, jiazheng.li\}@kcl.ac.uk}
 }
}

\begin{document}

\maketitle

\begin{abstract}
Good tutoring adapts to the individual: it tracks what a learner knows, notices \emph{why} they go wrong, and asks the next question that will help most.
Most deployed tutoring tools instead serve fixed item banks and treat a wrong answer as a single bit of signal.
We present \sys, an interactive, agentic tutor that supports an iterative
tutoring loop: the learner practises, and the system builds an
evidence-grounded memory of the learner's mastery and misconceptions.
This memory is updated as evidence accumulates and is used to generate the
next personalised question.
\sys has three components: (i) a persistent learner-state memory that updates per-topic mastery with a soft-evidence variant of Bayesian Knowledge Tracing, where a large language model acts as a continuous observation function; (ii) adaptive question generation that targets the learner's weakest topic and recurring misconceptions; and (iii) an evidence view that makes personalisation visible and testable through live progress visualisation and blind A/B comparison.
In blind A/B evaluation, questions conditioned on this memory are preferred over non-personalised ones $68$--$69\%$ of the time, and in persona simulations with hidden ground-truth mastery the agent's belief converges toward the learner's true mastery.
\end{abstract}

\section{Introduction}
\label{sec:intro}

Good teaching is not the scoring of answers as right or wrong. A long tradition in learning science treats genuine learning as \emph{conceptual change}: the learner gradually replaces a faulty understanding with a more accurate one \citep{piaget1952,smith1994misconceptions}, so a wrong answer is informative. It exposes the specific misconception the next round of teaching should address \citep{tatsuoka1983rulespace}, where earlier behaviourist accounts recorded only whether a response was correct \citep{thorndike1913}. Good instruction also meets each learner where they are, pitching the next task into the gap between what a student can do alone and what they can do with help \citep{vygotsky1978,bloom1984,vanlehn2011,li-etal-2023-distilling,li-etal-2024-calibrating,li-etal-2025-two, wong-etal-2026-questions}, which requires continuous diagnosis of the learner's present understanding \citep{black1998, li-etal-2026-towards}. 
The ideal tutor, in short, maintains a progressively refined account of each
learner's understanding and misconceptions and acts on it. However, a human
teacher who could continuously diagnose every student is unrealistic and
expensive.

Intelligent tutoring systems promise to scale this ideal, but existing deployed tools approximate it badly in three ways. \textbf{(i)}~\emph{Knowledge-tracing systems.} Classic adaptive tutors built on knowledge tracing \citep{corbett1994bkt,anderson1995cognitive,piech2015dkt} model knowledge \emph{coarsely}: every response is reduced to a correct/incorrect bit per skill, discarding the diagnostic content of a free-text answer. 
\textbf{(ii)}~\emph{LLM-based tutors.} Recent large language model tutors converse fluently and give high-quality feedback \citep{learnlm2024, kwon2024bipedpedagogicallyinformedtutoring, macina2023mathdial,class,li-etal-2025-automated,zhao-etal-2025-learnlens}, but are largely \emph{memoryless}.
\sys focuses on connecting answer-level diagnosis to persistent learner-state
memory and subsequent adaptive practice.
\textbf{(iii)}~\emph{Opaque adaptation.} In both families the adaptation 
is \emph{invisible}: the learner cannot inspect what the system believes about them \citep{bull2007smili,bodily2017dashboards}. 

\begin{figure*}[!tbh]
  \centering
  \includegraphics[width=\textwidth]{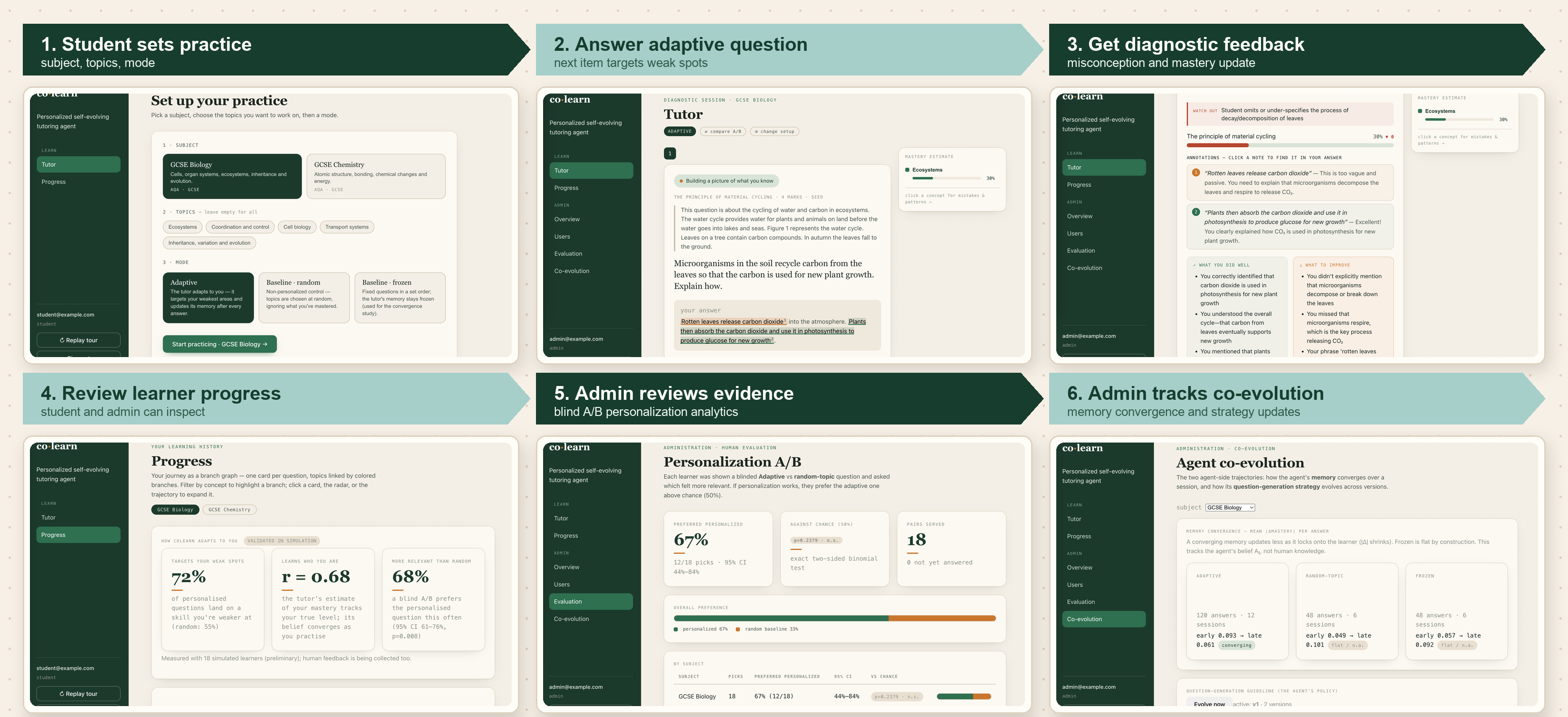}
  \caption{\sys demo workflow across learner and administrator views. The learner first configures a practice session, answers an adaptive question, and receives diagnostic feedback with answer-linked annotations and live mastery estimates. The progress and administrator views expose the same personalization evidence through learner progress, blind A/B analytics, and agent convergence diagnostics.}
  \label{fig:overview}
\end{figure*}

To close this gap and let more students benefit from personalised tutoring, we present \sys, an interactive tutor that operationalises the ideal tutor as a \emph{human--AI co-learning loop} and, crucially, makes its mastery and misconception estimates \emph{visible} and its personalisation \emph{testable}. The learner answers a question; the agent grades it, mines the misconception it reveals, updates a persistent per-topic mastery estimate, and uses that updated belief to generate the next, better-targeted question. The interface renders the agent's evolving belief in real time, so personalisation is something the user can see rather than take on trust, and a built-in blind A/B test
supports blinded comparison of adaptive and random-topic questions.
\sys is multi-subject (evaluated on Biology and Chemistry curricula), supports both multiple-choice and short-answer style questions, and deploys as a single-node container stack with a provider-selectable LLM backend. 
We use \emph{co-learning} in a deliberately modest sense: 
the learner practises the subject while the system refines its memory of the learner's
mastery and misconceptions. \sys surfaces how these estimates change and
stabilise during use, while the question strategy stays fixed
(\S\ref{sec:strategy}).
Our demo video is available at \td{\url{https://youtu.be/I1Lxg4ud3-Y}}.

\section{Overview of \sys}
\label{sec:system}

\begin{figure*}
\centering
\includegraphics[width=0.88\linewidth]{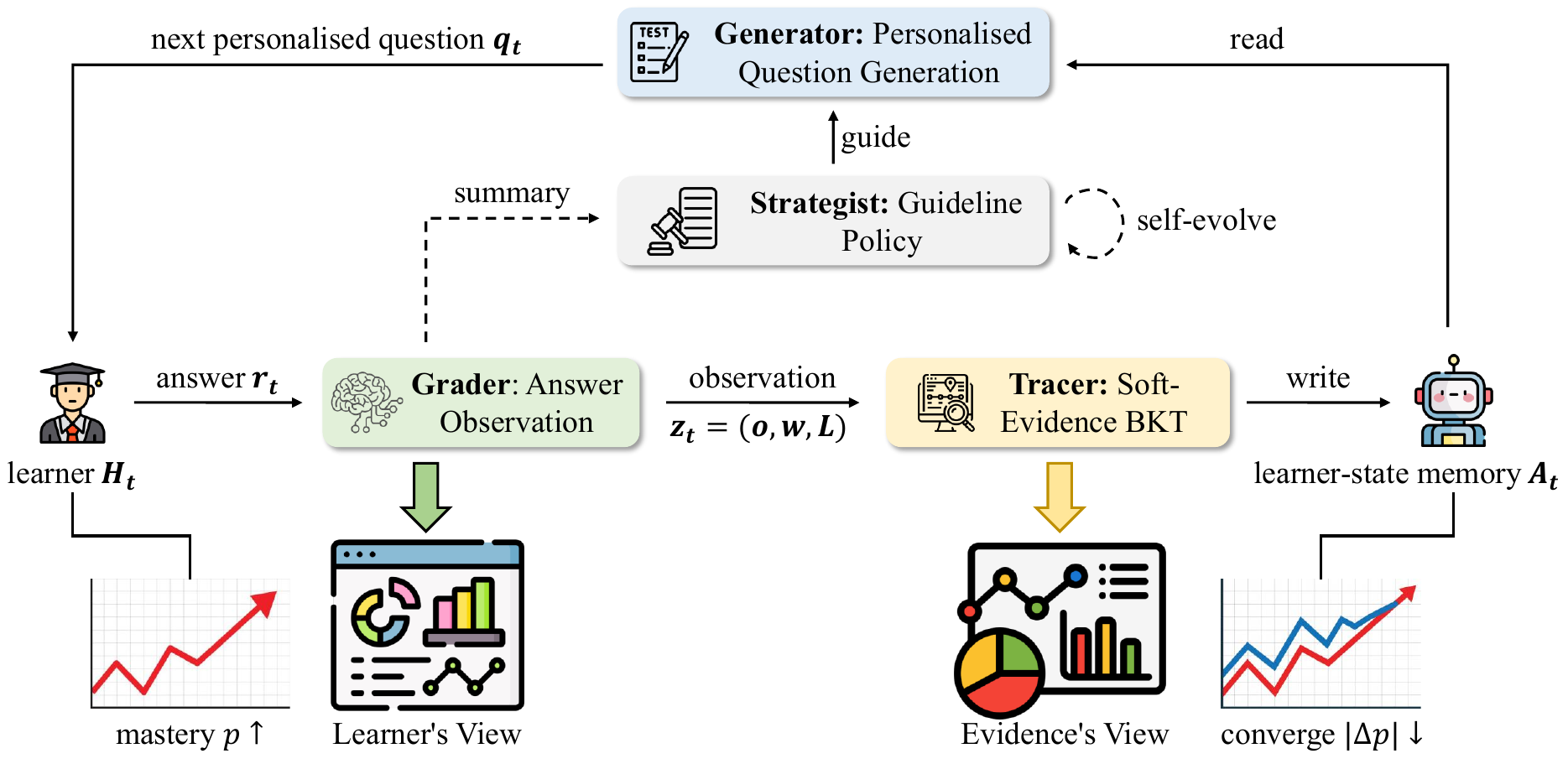}
\caption{The \sys co-learning loop. The learner answers; an LLM grades the answer into a continuous observation; soft-evidence BKT (Eq.~\ref{eq:bkt}) updates the agent's cross-session memory $A_t$; the personalised generator 
reads that memory and a fixed strategy guideline to write the next, better-targeted question; and the loop repeats (the dashed \emph{self-evolve} arrow is planned, not yet implemented). 
Two surfaces expose it: the \emph{learner view} (live feedback and mastery) and the \emph{evidence view} (progress dashboard + blind A/B).}
\label{fig:arch}
\end{figure*}

\sys is organised as \emph{two views of one loop} (Figure~\ref{fig:overview}). The \textbf{learner's view} (\S\ref{sec:learner}) is where practice happens: the learner answers a
question, receives feedback, and watches the agent's current mastery estimates fill in and update in real time. 
The \textbf{evidence view} (\S\ref{sec:evidence}) is where personalisation can be inspected through a progress dashboard and a blind A/B comparison of personalised and
non-personalised questions. 
Between the two sits the engine of the loop: a \textbf{persistent learner-state memory} that stores mastery estimates and mined misconception summaries (\S\ref{sec:memory}), and an \textbf{adaptive question generator} that acts on this memory (\S\ref{sec:quizgen}). Figure~\ref{fig:arch} sketches the full loop.

\subsection{Learner Interface}
\label{sec:learner}

A session begins with a short setup: the learner picks a \emph{subject}, the \emph{topics} to work on, and a \emph{mode} (\S\ref{sec:eval} explains the modes; the default is the adaptive system).

\sys then drives a streamed loop. It selects or generates the next item and streams its reasoning stages (``generating your question\ldots'') so the agent's work is visible. The learner answers (free text for short-answer items, or an option for multiple-choice), and the answer is graded with streamed stages (``marking your answer\ldots''). The feedback card shows a plain-language verdict, a ``watch out'' line surfacing the most salient misconception as friendly guidance, and, for free-text answers, the learner's own words re-rendered with span-level highlights linked to a mark-scheme breakdown.

After each answer, the targeted topic's mastery bar animates from its previous value to its new one, and a ``why this question'' chip names what the item was chosen to probe, making the personalisation explicit rather than implicit.

\subsection{Persistent Learner-State Memory}
\label{sec:memory}

To personalise tutoring across sessions, \sys maintains a persistent
learner-state memory containing topic-level mastery, recurring
misconception, and the mastery trajectory. Its contents are stored
in per-\emph{(learner, subject)} records rather than per-session logs, allowing
the agent to accumulate evidence over time and visualise how its belief evolves across sessions.

Formally, we model the interaction as a sequence of rounds: at round $t$ the agent poses a question $q_t$, the learner responds with $r_t$, and the agent returns feedback $f_t$. Each answered round drives an update $A_t\!\rightarrow\!A_{t+1}$ of the agent's belief about the learner, while the learner's true knowledge state $H_t$ remains latent. Concretely, $A_t$ holds a mastery estimate $p=P(\text{known})$ per topic, a de-duplicated list of mined misconceptions, and the mastery trajectory. \sys measures and visualises $A_t$; a rising mastery estimate is therefore evidence that the agent's belief has
changed,
not proof that the learner has mastered the skill (see the Limitations section).

\paragraph{Soft-Evidence Knowledge Tracing.}
Mastery is updated with a \textbf{soft-evidence} variant of Bayesian Knowledge Tracing \citep{corbett1994bkt}; extensions to standard BKT typically individualise its parameters or enrich its inputs \citep{yudelson2013individualized,piech2015dkt,abdelrahman2023survey,atk}. Standard BKT consumes a binary correct/incorrect observation. Instead, our LLM grader acts as a \emph{continuous observation function}: from an answer it emits graded evidence $o=\textit{mastery\_evidence}\in[0,1]$ together with a confidence $w=\textit{evidence\_strength}\in[0,1]$, and we feed these in without binarising at $0.5$. 
Given prior $p$ and BKT parameters $(p_{\text{slip}}, p_{\text{guess}}, p_{\text{transit}})$, we form the Bayesian posteriors for a correct and an incorrect observation,
\begin{align*}
p^{+} &= \frac{p\,(1-p_{\text{slip}})}{p\,(1-p_{\text{slip}})+(1-p)\,p_{\text{guess}}}, \\[2pt]
p^{-} &= \frac{p\,p_{\text{slip}}}{p\,p_{\text{slip}}+(1-p)\,(1-p_{\text{guess}})},
\end{align*}
blend them by $o$, and shrink toward the prior by the confidence $w$, giving the confidence-shrunk posterior $p_{\text{cond}} = p + w\,(o\,p^{+} + (1-o)\,p^{-} - p)$. We then apply a \emph{competence-gated} learning transition:
\begin{equation}
\label{eq:bkt}
p_{\text{next}} \;=\; p_{\text{cond}} \;+\; o\,(1-p_{\text{cond}})\,p_{\text{transit}}.
\end{equation}
Gating the learning term by $o$ ensures a clearly wrong answer cannot raise the mastery estimate; since standard BKT applies the transition after every attempt, the update is a \emph{variant} of BKT rather than a strict reduction (recovering standard BKT when $o{=}1$, $w{=}1$). The scheme is pure, unit-tested arithmetic with shared default parameters.

\paragraph{Misconception Memory.}
For seed items, candidate misconceptions are derived once from curriculum-aligned mark-scheme negative phrasing (e.g.\ ``do not accept\ldots'') and attached to the item, so the diagnostic targets come from the assessment material itself rather than a separately maintained catalogue. For \emph{short-answer questions}, the grader returns short misconception summaries with verbatim supporting quotes. It distinguishes a \emph{positive} label (the learner stated something wrong) from an \emph{absence} label (an omission or a too-vague answer), and reuses a candidate's canonical wording when one matches, so labels stay consistent across sessions; the summaries accumulate (de-duplicated) in memory. For \emph{multiple-choice items}, each distractor is tagged with the specific misconception it embodies, so \emph{choosing a distractor is itself the diagnostic label}, graded deterministically with no LLM call \citep{gierl2017distractors,wang2020diagnostic}.

\subsection{Adaptive Question Generation}
\label{sec:quizgen}

\textbf{Weakness-Targeted Generation.} 
To act on this memory, \sys generates the next item
to target their specific weaknesses \citep{kurdi2020systematic}. It ranks the learner's in-scope topics, targets the weakest, and selects up to two of its stored misconceptions to probe, optionally asking the LLM to link related misconceptions above a threshold. It then generates either a short-answer item or a multiple-choice item whose distractors are each bound to a known misconception. Generated items carry metadata (the targeted topic and misconceptions) that drives the visible ``why this question'' chip. When the learner has no in-scope history yet, curated seed items give a sensible cold start.

\textbf{Strategy Layer.}
\label{sec:strategy}
Beyond adapting its \emph{belief} about the learner, \sys is designed to adapt its \emph{strategy} that targets each individual learner's diagnosis based on the previous outcome: the question generator consults a natural-language guideline appended to its prompt (e.g.\ ``when targeting a stored misconception, prefer MCQ with one distractor per misconception''), so revising this text immediately changes how the next item is written, with no code change or redeployment. 
This is the hook for the agent-side half of co-learning, stated plainly: today the guideline is a \emph{fixed} text consulted on every generation; automatic revision from accumulated outcomes is planned but not yet implemented (see the Limitations section). The co-learning the live system demonstrates is therefore memory-side: the learner receives targeted practice while the agent updates its stored estimates from accumulated evidence.

\subsection{Evidence View}
\label{sec:evidence}

To make the agent's adaptation transparent, \sys provides a progress dashboard with a branch-graph timeline, concept-mastery radar, and mastery-over-time trajectory, 
allowing users to inspect how the mastery estimates and misconception summaries change across sessions.
\sys also includes a blind A/B interface to evaluate personalisation capability, presenting personalised and non-personalised questions in randomised, unlabelled positions and asking users which is more relevant. Results are aggregated into an analytics dashboard reporting the preference rate, a $95\%$ confidence interval, and a two-sided sign test against the $0.5$ chance rate.

\subsection{System Architecture}
\label{sec:impl}
\sys is a container-deployed web application. The frontend is a Vite\,+\,React\,+\,TypeScript single-page app. The backend is FastAPI with an async MongoDB store, JWT/bcrypt authentication, and SSE streaming for live generation and grading. The LLM backend is provider-selectable (AWS Bedrock with Claude Sonnet~4.5 as the live default; an OpenAI-compatible path is also supported). The whole stack starts with one command via Docker Compose (MongoDB, backend, a one-shot idempotent seeder, and an Nginx proxy serving the SPA and proxying the API). The backend ships with a test suite covering BKT, grading, generation, streaming, RBAC, and the A/B analytics.

\section{Evaluation}
\label{sec:eval}

We evaluate \sys along three dimensions: perceived relevance of personalised
versus non-personalised questions (\S\ref{sec:study1}), 
learner-state estimation accuracy and convergence under persona simulation (\S\ref{sec:eval-model}), 
and deployment responsiveness and cost (\S\ref{sec:study4}). 
\S\ref{sec:conditions} first defines the session conditions and metrics shared across these studies.

\subsection{Conditions and Metrics}
\label{sec:conditions}
\sys supports three session conditions that isolate \emph{personalisation} as the
variable of interest:
\begin{itemize}[leftmargin=*,itemsep=0pt,topsep=2pt,parsep=0pt]
\item \textbf{Adaptive}: the full system. Items are generated to target the learner's weakest topics, and memory is updated after every answer.
\item \textbf{Random-topic}: the non-personalised control. Items are drawn from topics chosen at random \emph{within the learner's selected scope} (mastery-blind), but memory is still updated. This isolates the effect of \emph{targeting} specifically, holding adaptivity constant.
\item \textbf{Frozen}: a memory-convergence control. Fixed items are served in order, and memory is read but not written, giving a flat agent-state baseline.
\end{itemize}

\subsection{Does Personalisation Produce More Relevant Questions?}
\label{sec:study1}
Using the built-in A/B tool, each comparison presents a personalised (Adaptive) question and a non-personalised (Random-topic) question in two blinded slots; slot order is randomised and provenance is hidden server-side. The rater answers the question ``which question feels more relevant to this learner right now?''. Picks are stored with their (hidden) provenance for later analysis; A/B items are not written back into the learner-state memory.

We recruited $18$ raters with a relevant background (e.g., they have studied and passed the curriculum). Each rater completed $12$ comparisons, yielding $M=18\times 12$ judgements. 
We report the \emph{preference rate} (fraction of comparisons in which the personalised item was chosen), its 95\% confidence interval, and a two-sided sign test against the chance rate of $0.5$. 

The personalised item is preferred $0.68$ of the time across Biology raters ($0.74$ and $0.73$ for mixed- and strong-ability learners) and $0.69$ across Chemistry raters ($p=0.008$, sign test over 18 raters). Preference is lower in the strong tier ($0.73$ Biology, $0.66$ Chemistry) than the mixed tier, consistent with targeting mattering less when there is little weakness to target; we do not test the six-rater tiers individually.

\begin{table}[t]
  \centering\small
  \resizebox{\columnwidth}{!}{%
  \begin{tabular}{lcccc}
    \toprule
    \textbf{Breakdown} & \textbf{$n$} & \textbf{Pref.\ rate} & \textbf{95\% CI} & \textbf{$p$} \\
    \midrule
    Biology (overall)       & 18 & 0.68 & [0.60, 0.77] & 0.008 \\
    \quad mixed learners  & 6  & 0.74 & [0.71, 0.77] &  \\
    \quad strong learners & 6  & 0.73 & [0.59, 0.88] &  \\
    Chemistry (overall)   & 18 & 0.69 & [0.60, 0.77] & 0.008 \\
    \quad mixed learners  & 6  & 0.77 & [0.61, 0.93] &  \\
    \quad strong learners & 6  & 0.66 & [0.52, 0.80] &  \\
    \bottomrule
  \end{tabular}}
  \caption{\textbf{Blind A/B preference for personalised over non-personalised questions.}
  \emph{Pref.\ rate} is the mean per-rater preference, \emph{95\% CI} is across raters,
  and \emph{$p$} is a two-sided sign test vs.\ $0.5$. The mixed/strong rows are learner ability tiers (not subjects); weak-tier learners are excluded because, being weak on every
  skill, they cannot discriminate targeting, so the tier rows sum to fewer than the
  \emph{all} row.}
  \label{tab:ab}
\end{table}

\begin{table}[t]
  \centering\small
    \begin{tabular}{lccc}
      \toprule
      \textbf{Metric} & \textbf{Adaptive} & \textbf{Random} & \textbf{Frozen} \\
      \midrule
      Weak-skill hit (warm)     & \textbf{0.72} & 0.55 & 0.57 \\
      \quad Mixed learners      & \textbf{0.89} & 0.57 & --   \\
      \quad Strong learners     & \textbf{0.28} & 0.08 & --   \\
      \addlinespace[2pt]
      \midrule
      Mastery MAE\,$\downarrow$ & \textbf{0.12} & 0.16 & 0.20 \\
      \bottomrule
    \end{tabular}
  \caption{\textbf{User-simulation} (hidden ground-truth mastery). Weak-skill hit rows
  are from an 18-persona warm-start sweep; the Mastery MAE row is from the 6-persona Claude Sonnet
  4.5-graded run. Weak-skill hit = fraction of served items on a truly-weak skill (higher = better
  targeting); Mastery MAE = final agent belief vs.\ true mastery (lower = better). Adaptive beats
  the Random-topic / Frozen controls on both. (Weak personas are weak on every skill, so they
  cannot discriminate targeting and are excluded from the per-tier rows.)}
  \label{tab:quality}
\end{table}

\begin{figure*}[t]
  \centering
  \includegraphics[width=\textwidth]{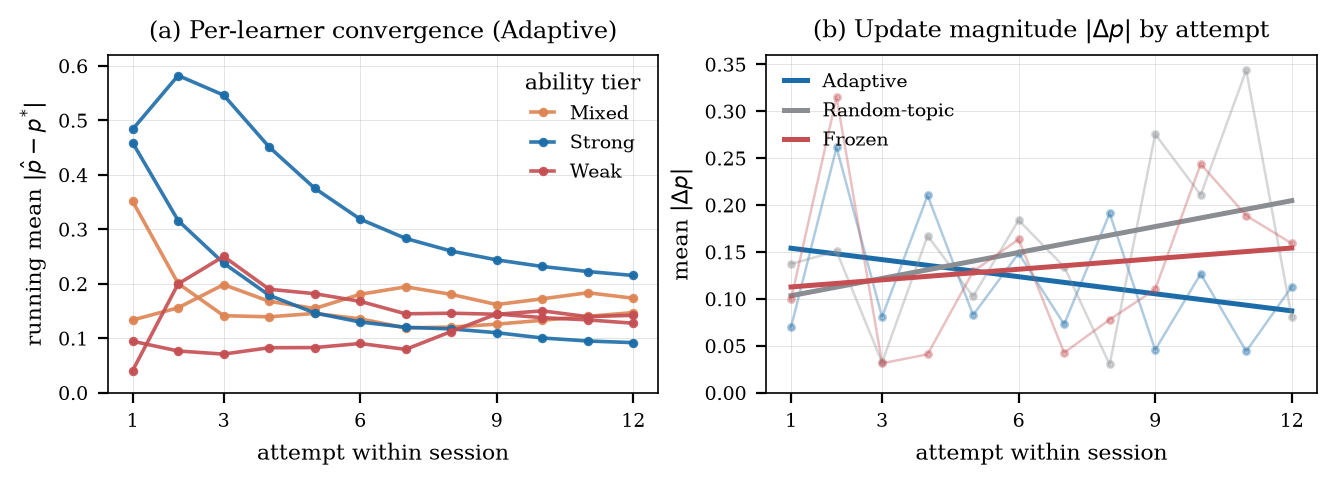}
  \caption{Memory convergence from the logged attempts (
   $6$ personas, $132$ rounds,
  live framework grader Claude Sonnet~4.5). \textbf{(a)} Per-learner
  running mean belief error $|\hat{p}-p^{*}|$ under \emph{Adaptive}: every
  learner's estimate settles toward its hidden true mastery, most visibly the two
  strong learners that start far off. \textbf{(b)} Mean update magnitude
  $|\Delta p|$ by attempt (thin = per-attempt mean, bold = linear trend): under
  \emph{Adaptive} it shrinks ($0.14\rightarrow0.10$) as the belief locks on, while
  the \emph{Random-topic} and \emph{Frozen} controls stay flat or drift upward.}
  \label{fig:convergence}
\end{figure*}

\subsection{Do Learner-State Estimates Track Ground Truth?}
\label{sec:eval-model}
Mastery prediction error and misconception P/R/F1 require ground truth, so we drive the loop with synthetic personas whose per-skill mastery and misconceptions are hidden from the system \citep{markel2023gpteach,tack2022aiteacher}: 
a database-free harness exercises the \emph{real} grading and BKT services, so answers flow through the live system's observation function.
On a run graded by the live framework model (Claude Sonnet 4.5; 6 personas, 132 rounds; 
the LLM grade serves as the observation, cf.\ LLM-as-judge reliability 
\citep{zheng2023judging}), %
per-answer \emph{mastery\_evidence} correlates with true mastery at Pearson $r=0.68$ pooled across tiers. This pooled figure is driven mainly by between-tier separation; however, mean evidence rises cleanly by ability tier ($0.25$ weak / $0.67$ mixed / $0.77$ strong), but the within-tier correlation is much weaker ($r\approx0.15$ / $0.48$ / $0.41$), and weakest for low-ability personas. The observation function is therefore least reliable for exactly the learners who most need accurate diagnosis (see Limitations).
Mined misconceptions match the personas' assigned ones at F1 $\approx0.56$.\footnote{Computed post hoc by token-overlap matching of mined labels against assigned misconceptions over all 132 rounds.} This F1 score evaluates the grader's recovery of persona misconception labels
from learner answers. It does not test whether a generated question elicits the
misconception named in its generation metadata.

\textbf{Accuracy and convergence.} A shrinking update magnitude shows only that the belief has \emph{stabilised}, not that it is correct, so accuracy against the hidden ground truth is the primary metric: final-belief mastery MAE is lowest under \emph{Adaptive} ($0.12$ vs.\ $0.16$/$0.20$; Table~\ref{tab:quality}), and the per-learner belief error $|\hat{p}-p^{*}|$ falls in step (Figure~\ref{fig:convergence}a). 
As a secondary aggregate signal, mean $|\Delta p|$ shrinks across the six \emph{Adaptive} persona sessions ($0.14\rightarrow0.10$) but stays flat or drifts upward under both controls (Figure~\ref{fig:convergence}b). Appendix~\ref{sec:appendix-trace} walks one live adaptive round end-to-end.

\subsection{Usability and Efficiency}
\label{sec:study4}
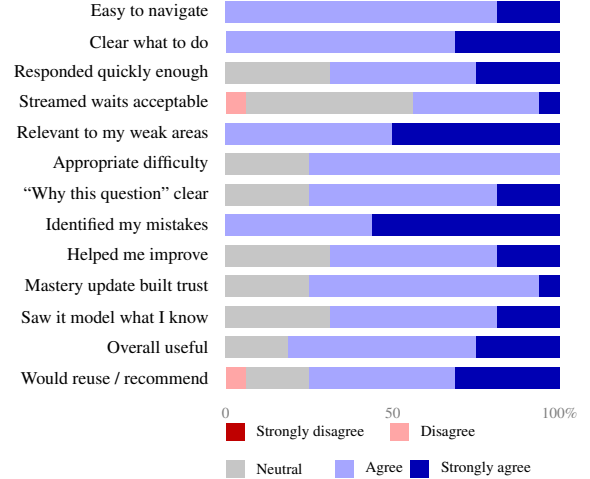
\begin{figure}[t]\centering
\resizebox{\columnwidth}{!}{%
\begin{tikzpicture}[font=\scriptsize]
\node[anchor=east] at (-0.1,0.15) {Easy to navigate};
\fill[blue!35] (0.000,0.00) rectangle (3.737,0.30);
\fill[blue!70!black] (3.737,0.00) rectangle (4.600,0.30);
\node[anchor=east] at (-0.1,-0.27) {Clear what to do};
\fill[blue!35] (0.000,-0.42) rectangle (3.162,-0.12);
\fill[blue!70!black] (3.162,-0.42) rectangle (4.600,-0.12);
\node[anchor=east] at (-0.1,-0.69) {Responded quickly enough};
\fill[gray!45] (0.000,-0.84) rectangle (1.438,-0.54);
\fill[blue!35] (1.438,-0.84) rectangle (3.450,-0.54);
\fill[blue!70!black] (3.450,-0.84) rectangle (4.600,-0.54);
\node[anchor=east] at (-0.1,-1.11) {Streamed waits acceptable};
\fill[red!35] (0.000,-1.26) rectangle (0.287,-0.96);
\fill[gray!45] (0.287,-1.26) rectangle (2.587,-0.96);
\fill[blue!35] (2.587,-1.26) rectangle (4.312,-0.96);
\fill[blue!70!black] (4.312,-1.26) rectangle (4.600,-0.96);
\node[anchor=east] at (-0.1,-1.53) {Relevant to my weak areas};
\fill[blue!35] (0.000,-1.68) rectangle (2.300,-1.38);
\fill[blue!70!black] (2.300,-1.68) rectangle (4.600,-1.38);
\node[anchor=east] at (-0.1,-1.95) {Appropriate difficulty};
\fill[gray!45] (0.000,-2.10) rectangle (1.150,-1.80);
\fill[blue!35] (1.150,-2.10) rectangle (4.600,-1.80);
\node[anchor=east] at (-0.1,-2.37) {``Why this question'' clear};
\fill[gray!45] (0.000,-2.52) rectangle (1.150,-2.22);
\fill[blue!35] (1.150,-2.52) rectangle (3.737,-2.22);
\fill[blue!70!black] (3.737,-2.52) rectangle (4.600,-2.22);
\node[anchor=east] at (-0.1,-2.79) {Identified my mistakes};
\fill[blue!35] (0.000,-2.94) rectangle (2.012,-2.64);
\fill[blue!70!black] (2.012,-2.94) rectangle (4.600,-2.64);
\node[anchor=east] at (-0.1,-3.21) {Helped me improve};
\fill[gray!45] (0.000,-3.36) rectangle (1.438,-3.06);
\fill[blue!35] (1.438,-3.36) rectangle (3.737,-3.06);
\fill[blue!70!black] (3.737,-3.36) rectangle (4.600,-3.06);
\node[anchor=east] at (-0.1,-3.63) {Mastery update built trust};
\fill[gray!45] (0.000,-3.78) rectangle (1.150,-3.48);
\fill[blue!35] (1.150,-3.78) rectangle (4.312,-3.48);
\fill[blue!70!black] (4.312,-3.78) rectangle (4.600,-3.48);
\node[anchor=east] at (-0.1,-4.05) {Saw it model what I know};
\fill[gray!45] (0.000,-4.20) rectangle (1.438,-3.90);
\fill[blue!35] (1.438,-4.20) rectangle (3.737,-3.90);
\fill[blue!70!black] (3.737,-4.20) rectangle (4.600,-3.90);
\node[anchor=east] at (-0.1,-4.47) {Overall useful};
\fill[gray!45] (0.000,-4.62) rectangle (0.862,-4.32);
\fill[blue!35] (0.862,-4.62) rectangle (3.450,-4.32);
\fill[blue!70!black] (3.450,-4.62) rectangle (4.600,-4.32);
\node[anchor=east] at (-0.1,-4.89) {Would reuse / recommend};
\fill[red!35] (0.000,-5.04) rectangle (0.287,-4.74);
\fill[gray!45] (0.287,-5.04) rectangle (1.150,-4.74);
\fill[blue!35] (1.150,-5.04) rectangle (3.162,-4.74);
\fill[blue!70!black] (3.162,-5.04) rectangle (4.600,-4.74);
\node[anchor=north,font=\tiny,gray] at (0.00,-5.16) {0};
\node[anchor=north,font=\tiny,gray] at (2.30,-5.16) {50};
\node[anchor=north,font=\tiny,gray] at (4.60,-5.16) {100\%};
\fill[red!75!black] (0.00,-5.76) rectangle (0.26,-5.52);
\node[anchor=west,font=\tiny] at (0.29,-5.64) {Strongly disagree};
\fill[red!35] (2.26,-5.76) rectangle (2.52,-5.52);
\node[anchor=west,font=\tiny] at (2.55,-5.64) {Disagree};
\fill[gray!45] (0.00,-6.26) rectangle (0.26,-6.02);
\node[anchor=west,font=\tiny] at (0.29,-6.14) {Neutral};
\fill[blue!35] (1.50,-6.26) rectangle (1.76,-6.02);
\node[anchor=west,font=\tiny] at (1.79,-6.14) {Agree};
\fill[blue!70!black] (2.54,-6.26) rectangle (2.80,-6.02);
\node[anchor=west,font=\tiny] at (2.83,-6.14) {Strongly agree};
\end{tikzpicture}}
\caption{Per-item response distribution of the 1--5 Likert questionnaire over 16 raters. Blue = agree, grey = neutral, red = disagree. No item drew a strongly-disagree response; the weakest is the streamed-wait item (Q4), echoing the measured latency (\S\ref{sec:efficiency}).}
\label{fig:survey}
\end{figure}

To evaluate the learner-facing experience, we asked $16$ human raters to answer a 1--5 Likert questionnaire of 13 items across six dimensions after a session (full instrument in Appendix~\ref{sec:appendix-survey}). This questionnaire are
designed to capture perceived relevance and usefulness. It does not evaluate
whether generated questions reliably elicit their intended misconceptions. Figure~\ref{fig:survey} shows the per-item distribution; the overall mean is $4.0/5$, no dimension below ${\approx}3.7$, and $75\%$ ($12/16$) would reuse. The
lowest-rated dimension is \emph{Responsiveness} ($3.7$), consistent with the measured short-answer
latency (\S\ref{sec:efficiency}); researchers were the most critical (mean $3.8$) and students the
most positive ($4.4$).

\textbf{Efficiency.}
\begin{table}[!t]
  \centering\small
  \resizebox{\columnwidth}{!}{%
  \begin{tabular}{lccc}
    \toprule
    \textbf{Round type} & \textbf{Next/gen (s)} & \textbf{Answer (s)} & \textbf{Cost (\$/rnd)} \\
    \midrule
    Short-answer & 12.1 & 20.2 & $\approx$0.015 \\
    Multiple-choice & 7.6 & 0.01 & $\approx$0.005 \\
    \bottomrule
  \end{tabular}}
  \caption{Efficiency on the live backend (Bedrock Claude Sonnet 4.5). Short-answer $\approx$32\,s/round (grade+feedback dominate); MCQ $\approx$7.6\,s (deterministic answer). Cost is estimated from token usage.}
  \label{tab:cost}
\end{table}
\label{sec:efficiency}
Deployability hinges on cost and latency. A short-answer round uses two LLM calls
(grading and feedback) plus generation of the next item; a multiple-choice round
is graded and explained \emph{deterministically} with no LLM call. Table~\ref{tab:cost}
reports mean latency measured on the live single-node deployment (AWS Bedrock, Claude
Sonnet 4.5). The grader/feedback pass dominates short-answer latency; multiple-choice
rounds are $\approx4\times$ faster because only the next-item generation calls the LLM. A
faster model (e.g.\ Claude Haiku) is the obvious lever for a booth setting.

\section{Conclusion}
\label{sec:conclusion}

\sys demonstrates how an LLM tutor can maintain a persistent 
learner-state memory, use that memory to generate personalised questions, 
and expose the resulting adaptation through live visualisation and built-in A/B evaluation. 
Its soft-evidence knowledge-tracing memory converges on the learner across sessions, its generator targets the learner's weakest topics and misconceptions, and its evidence view makes the personalisation testable by anyone. 
By connecting learner-state estimation,, question generation, and transparent evidence surfaces in one deployable, container-deployed research prototype, \sys provides a platform for the harder question ahead: whether visible, converging personalisation produces durable, transferable learning.

\section*{Acknowledgments}
This work was supported in part by the UK Engineering and Physical Sciences Research Council (EPSRC) through a Turing AI Fellowship (grant no. EP/V020579/1, EP/V020579/2) and the Prosperity Partnership scheme (grant no. UKRI566).

\section*{Limitations}
\label{sec:limitations}
The agent's mastery estimate is its belief $A_t$, not a direct measurement of
human knowledge $H_t$; we therefore do not claim learning gains, and our A/B study
measures \emph{perceived} relevance rather than learning outcomes. BKT parameters are
shared priors, not fit per skill, so absolute mastery values are indicative rather
than calibrated.
The LLM observation function is also least reliable for low-performing learners: within-tier, per-answer evidence correlates with true mastery at only $r\approx0.15$ for the weak tier, versus $0.48$/$0.41$ for mixed/strong (the higher pooled $r=0.68$ reflects between-tier separation, not within-tier accuracy). These are precisely the learners who most need accurate diagnosis; improving grading fidelity on weak answers is a priority for future work.
Our evaluation does not establish that generated questions elicit their intended misconceptions, nor does it independently assess factual correctness, curriculum fit, answerability, or distractor quality. These properties require expert annotation and learner-response studies.
Automatic strategy self-evolution and the co-evolution diagnostics
(convergence/transfer estimation) 
are designed but not yet implemented: during use the strategy guideline is fixed, so only the agent's memory adapts.
Far-transfer learning, the strongest test of
genuine competence, is out of scope here and left to future work.
Due to differences in experimental setup and evaluation protocols, we do not
make direct comparisons with other LLM-based tutoring systems.

\section*{Ethics Statement}
\label{sec:ethics}
\sys involves human participants (the A/B and survey raters) and learner data. Accounts are admin-provisioned with no public signup, learner data is stored locally in the single-node deployment, and the system can run fully offline with a deterministic LLM for privacy-sensitive settings. Misconception summaries are surfaced to learners as supportive guidance rather than verbatim labels.

\bibliography{custom}

@book{piaget1952,
  title={The Origins of Intelligence in Children},
  author={Piaget, J.},
  isbn={9780823682072},
  lccn={52014807},
  series={International Universities Press paperback library},
  url={https://books.google.com.sg/books?id=H7MkAQAAMAAJ},
  year={1952},
  publisher={International Universities Press}
}

@book{thorndike1913,
  title={The psychology of learning},
  author={Thorndike, E.L.},
  series={Educational Psychology},
  url={https://books.google.com.sg/books?id=zLkBAAAAYAAJ},
  year={1913},
  publisher={Teachers College, Columbia University}
}

@book{vygotsky1978,
 ISBN = {9780674576285},
 URL = {http://www.jstor.org/stable/j.ctvjf9vz4},
 author = {L. S. Vygotsky},
 publisher = {Harvard University Press},
 title = {Mind in Society: Development of Higher Psychological Processes},
 urldate = {2026-07-08},
 year = {1978}
}

@article{bloom1984,
author = {Benjamin S. Bloom},
title ={The 2 Sigma Problem: The Search for Methods of Group Instruction as Effective as One-to-One Tutoring},
journal = {Educational Researcher},
volume = {13},
number = {6},
pages = {4-16},
year = {1984},
doi = {10.3102/0013189X013006004},
URL = {https://doi.org/10.3102/0013189X013006004},
eprint = {https://doi.org/10.3102/0013189X013006004}
}

@article{black1998,
author = {Black, Paul and Wiliam, Dylan},
year = {1998},
month = {03},
pages = {7-74},
title = {Assessment and Classroom Learning},
volume = {5},
journal = {Assessment in Education},
doi = {10.1080/0969595980050102}
}

@article{smith1994misconceptions,
author = {Smith III, John P. and Andrea A. diSessa and Jeremy Roschelle},
title = {Misconceptions Reconceived: A Constructivist Analysis of Knowledge in Transition},
journal = {Journal of the Learning Sciences},
volume = {3},
number = {2},
pages = {115--163},
year = {1994},
publisher = {Routledge},
doi = {10.1207/s15327809jls0302\_1},
URL = {https://doi.org/10.1207/s15327809jls0302_1},
eprint = {https://doi.org/10.1207/s15327809jls0302_1}
}

@article{vanlehn2011,
author = {Kurt VanLEHN},
title = {The Relative Effectiveness of Human Tutoring, Intelligent Tutoring Systems, and Other Tutoring Systems},
journal = {Educational Psychologist},
volume = {46},
number = {4},
pages = {197--221},
year = {2011},
publisher = {Routledge},
doi = {10.1080/00461520.2011.611369},
URL = {https://doi.org/10.1080/00461520.2011.611369},
eprint = {https://doi.org/10.1080/00461520.2011.611369}
}

@article{corbett1994bkt,
  author = {Corbett, Albert T. and Anderson, John R.},
  title = {Knowledge tracing: Modeling the acquisition of procedural knowledge},
  journal = {User Modeling and User-Adapted Interaction},
  year = {1994},
  volume = {4},
  number = {4},
  pages = {253--278},
  doi = {10.1007/BF01099821},
  url = {https://doi.org/10.1007/BF01099821},
  isbn = {1573-1391}
}

@article{anderson1995cognitive,
author = {John R. Anderson and Albert T. Corbett and Kenneth R. Koedinger and Ray. Pelletier},
title = {Cognitive Tutors: Lessons Learned},
journal = {Journal of the Learning Sciences},
volume = {4},
number = {2},
pages = {167--207},
year = {1995},
publisher = {Routledge},
doi = {10.1207/s15327809jls0402\_2},
URL = {https://doi.org/10.1207/s15327809jls0402_2},
eprint = {https://doi.org/10.1207/s15327809jls0402_2}
}

@inproceedings{piech2015dkt,
 author = {Piech, Chris and Bassen, Jonathan and Huang, Jonathan and Ganguli, Surya and Sahami, Mehran and Guibas, Leonidas and Sohl-Dickstein, Jascha},
 booktitle = {Advances in Neural Information Processing Systems},
 editor = {C. Cortes and N. Lawrence and D. Lee and M. Sugiyama and R. Garnett},
 pages = {},
 publisher = {Curran Associates, Inc.},
 title = {Deep Knowledge Tracing},
 url = {https://proceedings.neurips.cc/paper_files/paper/2015/file/bac9162b47c56fc8a4d2a519803d51b3-Paper.pdf},
 volume = {28},
 year = {2015}
}

@InProceedings{yudelson2013individualized,
author="Yudelson, Michael V.
and Koedinger, Kenneth R.
and Gordon, Geoffrey J.",
editor="Lane, H. Chad
and Yacef, Kalina
and Mostow, Jack
and Pavlik, Philip",
title="Individualized Bayesian Knowledge Tracing Models",
booktitle="Artificial Intelligence in Education",
year="2013",
publisher="Springer Berlin Heidelberg",
address="Berlin, Heidelberg",
pages="171--180",
isbn="978-3-642-39112-5"
}

@article{abdelrahman2023survey,
author = {Abdelrahman, Ghodai and Wang, Qing and Nunes, Bernardo},
title = {Knowledge Tracing: A Survey},
year = {2023},
issue_date = {November 2023},
publisher = {Association for Computing Machinery},
address = {New York, NY, USA},
volume = {55},
number = {11},
issn = {0360-0300},
url = {https://doi.org/10.1145/3569576},
doi = {10.1145/3569576},
journal = {ACM Comput. Surv.},
month = feb,
articleno = {224},
numpages = {37}
}

@article{bull2007smili,
author = {Susan Bull and Judy Kay},
title ={Student Models that Invite the Learner In: The SMILI:() Open Learner Modelling Framework},
journal = {International Journal of Artificial Intelligence in Education},
volume = {17},
number = {2},
pages = {89-120},
year = {2007},
doi = {10.3233/IRG-2007-17(2)02},
URL = {https://journals.sagepub.com/doi/abs/10.3233/IRG-2007-17%282%2902},
eprint = {https://journals.sagepub.com/doi/pdf/10.3233/IRG-2007-17%282%2902}
}

@ARTICLE{bodily2017dashboards,
  author={Bodily, Robert and Verbert, Katrien},
  journal={IEEE Transactions on Learning Technologies}, 
  title={Review of Research on Student-Facing Learning Analytics Dashboards and Educational Recommender Systems}, 
  year={2017},
  volume={10},
  number={4},
  pages={405-418},
  doi={10.1109/TLT.2017.2740172}}

@misc{learnlm2024,
      title={LearnLM: Improving Gemini for Learning}, 
      author={{LearnLM Team} and Abhinit Modi and Aditya Srikanth Veerubhotla and Aliya Rysbek and Andrea Huber and Brett Wiltshire and Brian Veprek and Daniel Gillick and Daniel Kasenberg and Derek Ahmed and Irina Jurenka and James Cohan and Jennifer She and Julia Wilkowski and Kaiz Alarakyia and Kevin R. McKee and Lisa Wang and Markus Kunesch and Mike Schaekermann and Miruna Pîslar and Nikhil Joshi and Parsa Mahmoudieh and Paul Jhun and Sara Wiltberger and Shakir Mohamed and Shashank Agarwal and Shubham Milind Phal and Sun Jae Lee and Theofilos Strinopoulos and Wei-Jen Ko and Amy Wang and Ankit Anand and Avishkar Bhoopchand and Dan Wild and Divya Pandya and Filip Bar and Garth Graham and Holger Winnemoeller and Mahvish Nagda and Prateek Kolhar and Renee Schneider and Shaojian Zhu and Stephanie Chan and Steve Yadlowsky and Viknesh Sounderajah and Yannis Assael},
      year={2025},
      eprint={2412.16429},
      archivePrefix={arXiv},
      primaryClass={cs.CY},
      url={https://arxiv.org/abs/2412.16429}, 
}

@inproceedings{macina2023mathdial,
    title = "{M}ath{D}ial: A Dialogue Tutoring Dataset with Rich Pedagogical Properties Grounded in Math Reasoning Problems",
    author = "Macina, Jakub  and
      Daheim, Nico  and
      Chowdhury, Sankalan  and
      Sinha, Tanmay  and
      Kapur, Manu  and
      Gurevych, Iryna  and
      Sachan, Mrinmaya",
    editor = "Bouamor, Houda  and
      Pino, Juan  and
      Bali, Kalika",
    booktitle = "Findings of the Association for Computational Linguistics: EMNLP 2023",
    month = dec,
    year = "2023",
    address = "Singapore",
    publisher = "Association for Computational Linguistics",
    url = "https://aclanthology.org/2023.findings-emnlp.372/",
    doi = "10.18653/v1/2023.findings-emnlp.372",
    pages = "5602--5621"
}

@inproceedings{markel2023gpteach,
author = {Markel, Julia M. and Opferman, Steven G. and Landay, James A. and Piech, Chris},
title = {GPTeach: Interactive TA Training with GPT-based Students},
year = {2023},
isbn = {9798400700255},
publisher = {Association for Computing Machinery},
address = {New York, NY, USA},
url = {https://doi.org/10.1145/3573051.3593393},
doi = {10.1145/3573051.3593393},
booktitle = {Proceedings of the Tenth ACM Conference on Learning @ Scale},
pages = {226–236},
numpages = {11},
location = {Copenhagen, Denmark},
series = {L@S '23}
}

@misc{tack2022aiteacher,
      title={The AI Teacher Test: Measuring the Pedagogical Ability of Blender and GPT-3 in Educational Dialogues}, 
      author={Anaïs Tack and Chris Piech},
      year={2022},
      eprint={2205.07540},
      archivePrefix={arXiv},
      primaryClass={cs.CL},
      url={https://arxiv.org/abs/2205.07540}, 
}

@misc{kwon2024bipedpedagogicallyinformedtutoring,
      title={BIPED: Pedagogically Informed Tutoring System for ESL Education}, 
      author={Soonwoo Kwon and Sojung Kim and Minju Park and Seunghyun Lee and Kyuseok Kim},
      year={2024},
      eprint={2406.03486},
      archivePrefix={arXiv},
      primaryClass={cs.CL},
      url={https://arxiv.org/abs/2406.03486}, 
}

@article{kurdi2020systematic,
title = {A Systematic Review of Automatic Question Generation for Educational Purposes},
journal = {International Journal of Artificial Intelligence in Education},
volume = {30},
number = {1},
pages = {121-204},
year = {2020},
issn = {1560-4292},
doi = {https://doi.org/10.1007/s40593-019-00186-y},
url = {https://www.sciencedirect.com/science/article/pii/S1560429226006967},
author = {Ghader Kurdi and Jared Leo and Bijan Parsia and Uli Sattler and Salam Al-Emari}
}

@article{gierl2017distractors,
author = {Mark J. Gierl and Okan Bulut and Qi Guo and Xinxin Zhang},
title ={Developing, Analyzing, and Using Distractors for Multiple-Choice Tests in Education: A Comprehensive Review},
journal = {Review of Educational Research},
volume = {87},
number = {6},
pages = {1082-1116},
year = {2017},
doi = {10.3102/0034654317726529},
URL = {https://doi.org/10.3102/0034654317726529},
eprint = {https://doi.org/10.3102/0034654317726529}
}

@misc{wang2020diagnostic,
      title={Instructions and Guide for Diagnostic Questions: The NeurIPS 2020 Education Challenge}, 
      author={Zichao Wang and Angus Lamb and Evgeny Saveliev and Pashmina Cameron and Yordan Zaykov and José Miguel Hernández-Lobato and Richard E. Turner and Richard G. Baraniuk and Craig Barton and Simon Peyton Jones and Simon Woodhead and Cheng Zhang},
      year={2021},
      eprint={2007.12061},
      archivePrefix={arXiv},
      primaryClass={cs.CY},
      url={https://arxiv.org/abs/2007.12061}, 
}

@misc{zheng2023judging,
      title={Judging LLM-as-a-Judge with MT-Bench and Chatbot Arena}, 
      author={Lianmin Zheng and Wei-Lin Chiang and Ying Sheng and Siyuan Zhuang and Zhanghao Wu and Yonghao Zhuang and Zi Lin and Zhuohan Li and Dacheng Li and Eric P. Xing and Hao Zhang and Joseph E. Gonzalez and Ion Stoica},
      year={2023},
      eprint={2306.05685},
      archivePrefix={arXiv},
      primaryClass={cs.CL},
      url={https://arxiv.org/abs/2306.05685}, 
}

@misc{class,
      title={CLASS: A Design Framework for building Intelligent Tutoring Systems based on Learning Science principles}, 
      author={Shashank Sonkar and Naiming Liu and Debshila Basu Mallick and Richard G. Baraniuk},
      year={2023},
      eprint={2305.13272},
      archivePrefix={arXiv},
      primaryClass={cs.CL},
      url={https://arxiv.org/abs/2305.13272}, 
}

@misc{atk,
      title={Context-Aware Attentive Knowledge Tracing}, 
      author={Aritra Ghosh and Neil Heffernan and Andrew S. Lan},
      year={2020},
      eprint={2007.12324},
      archivePrefix={arXiv},
      primaryClass={cs.LG},
      url={https://arxiv.org/abs/2007.12324}, 
}

@article{tatsuoka1983rulespace,
author = {Tatsuoka, KIKUMI K.},
title = {RULE SPACE: AN APPROACH FOR DEALING WITH MISCONCEPTIONS BASED ON ITEM RESPONSE THEORY},
journal = {Journal of Educational Measurement},
volume = {20},
number = {4},
pages = {345-354},
doi = {https://doi.org/10.1111/j.1745-3984.1983.tb00212.x},
url = {https://onlinelibrary.wiley.com/doi/abs/10.1111/j.1745-3984.1983.tb00212.x},
eprint = {https://onlinelibrary.wiley.com/doi/pdf/10.1111/j.1745-3984.1983.tb00212.x},
year = {1983}
}

@inproceedings{li-etal-2023-distilling,
    title = "Distilling {C}hat{GPT} for Explainable Automated Student Answer Assessment",
    author = "Li, Jiazheng  and
      Gui, Lin  and
      Zhou, Yuxiang  and
      West, David  and
      Aloisi, Cesare  and
      He, Yulan",
    editor = "Bouamor, Houda  and
      Pino, Juan  and
      Bali, Kalika",
    booktitle = "Findings of the Association for Computational Linguistics: EMNLP 2023",
    month = dec,
    year = "2023",
    address = "Singapore",
    publisher = "Association for Computational Linguistics",
    url = "https://aclanthology.org/2023.findings-emnlp.399/",
    doi = "10.18653/v1/2023.findings-emnlp.399",
    pages = "6007--6026"
}

@inproceedings{li-etal-2024-calibrating,
    title = "Calibrating {LLM}s with Preference Optimization on Thought Trees for Generating Rationale in Science Question Scoring",
    author = "Li, Jiazheng  and
      Xu, Hainiu  and
      Sun, Zhaoyue  and
      Zhou, Yuxiang  and
      West, David  and
      Aloisi, Cesare  and
      He, Yulan",
    editor = "Al-Onaizan, Yaser  and
      Bansal, Mohit  and
      Chen, Yun-Nung",
    booktitle = "Findings of the Association for Computational Linguistics: EMNLP 2024",
    month = nov,
    year = "2024",
    address = "Miami, Florida, USA",
    publisher = "Association for Computational Linguistics",
    url = "https://aclanthology.org/2024.findings-emnlp.313/",
    doi = "10.18653/v1/2024.findings-emnlp.313",
    pages = "5452--5479"
}

@inproceedings{li-etal-2025-two,
    title = "Two Heads Are Better Than One: Dual-Model Verbal Reflection at Inference-Time",
    author = "Li, Jiazheng  and
      Zhou, Yuxiang  and
      Lu, Junru  and
      Tyen, Gladys  and
      Gui, Lin  and
      Aloisi, Cesare  and
      He, Yulan",
    editor = "Christodoulopoulos, Christos  and
      Chakraborty, Tanmoy  and
      Rose, Carolyn  and
      Peng, Violet",
    booktitle = "Proceedings of the 2025 Conference on Empirical Methods in Natural Language Processing",
    month = nov,
    year = "2025",
    address = "Suzhou, China",
    publisher = "Association for Computational Linguistics",
    url = "https://aclanthology.org/2025.emnlp-main.155/",
    doi = "10.18653/v1/2025.emnlp-main.155",
    pages = "3119--3140",
    ISBN = "979-8-89176-332-6"
}

@inproceedings{zhao-etal-2025-learnlens,
    title = "{L}earn{L}ens: {LLM}-Enabled Personalised, Curriculum-Grounded Feedback with Educators in the Loop",
    author = "Zhao, Runcong  and
      Bobrov, Artem  and
      Li, Jiazheng  and
      Aloisi, Cesare  and
      He, Yulan",
    editor = {Habernal, Ivan  and
      Schulam, Peter  and
      Tiedemann, J{\"o}rg},
    booktitle = "Proceedings of the 2025 Conference on Empirical Methods in Natural Language Processing: System Demonstrations",
    month = nov,
    year = "2025",
    address = "Suzhou, China",
    publisher = "Association for Computational Linguistics",
    url = "https://aclanthology.org/2025.emnlp-demos.45/",
    doi = "10.18653/v1/2025.emnlp-demos.45",
    pages = "625--633",
    ISBN = "979-8-89176-334-0"
}

@misc{li-etal-2025-automated,
      title={An Automated Explainable Educational Assessment System Built on LLMs}, 
      author={Jiazheng Li and Artem Bobrov and David West and Cesare Aloisi and Yulan He},
      year={2024},
      eprint={2412.13381},
      archivePrefix={arXiv},
      primaryClass={cs.CL},
      url={https://arxiv.org/abs/2412.13381}, 
}

@inproceedings{li-etal-2026-towards,
    title = "Towards Self-Improving Error Diagnosis in Multi-Agent Systems",
    author = "Li, Jiazheng  and
      Yilmaz, Emine  and
      Chen, Bei  and
      Le, Thu",
    editor = "Liakata, Maria  and
      Moreira, Viviane P.  and
      Zhang, Jiajun  and
      Jurgens, David",
    booktitle = "Findings of the {A}ssociation for {C}omputational {L}inguistics: {ACL} 2026",
    month = jul,
    year = "2026",
    address = "San Diego, California, United States",
    publisher = "Association for Computational Linguistics",
    url = "https://aclanthology.org/2026.findings-acl.98/",
    doi = "10.18653/v1/2026.findings-acl.98",
    pages = "2063--2077",
    ISBN = "979-8-89176-395-1"
}

@inproceedings{wong-etal-2026-questions,
    title = "From Questions to Assessment Tuples: A Multi-Agent Framework with Bloom-Specialized Agents and Automated Verification",
    author = "Wong, Gee-Lyle  and
      Zhao, Runcong  and
      He, Yulan  and
      Li, Jiazheng",
    editor = "Kochmar, Ekaterina  and
      Alhafni, Bashar  and
      Bann{\`o}, Stefano  and
      Bexte, Marie  and
      Burstein, Jill  and
      Horbach, Andrea  and
      Laarmann-Quante, Ronja  and
      Tack, Anais  and
      Yaneva, Victoria  and
      Yuan, Zheng",
    booktitle = "Proceedings of the 21st Workshop on Innovative Use of {NLP} for Building Educational Applications ({BEA} 2026)",
    month = jul,
    year = "2026",
    address = "San Diego, California, USA",
    publisher = "Association for Computational Linguistics",
    url = "https://aclanthology.org/2026.bea-1.22/",
    doi = "10.18653/v1/2026.bea-1.22",
    pages = "292--335",
    ISBN = "979-8-89176-409-5"
}

\appendix
\section{User-study questionnaire}
\label{sec:appendix-survey}
The instrument used in the user study (\S\ref{sec:study4}). Each statement is rated 1--5 (1 = strongly
disagree, 5 = strongly agree); Q13 is the adoption item. Raters span a range of backgrounds
(teacher, researcher, student, parent) and complete the items after a session.
 
\begin{itemize}\itemsep1pt
\item[] \emph{Usability.} \\Q1 The interface was easy to navigate.\\Q2 At each step it was
clear what I was meant to do.
\item[] \emph{Responsiveness.}\\Q3 The system responded quickly enough to keep me engaged.\\
Q4 The streamed ``thinking'' stages made any wait feel acceptable.
\item[] \emph{Question quality.}\\Q5 The questions were relevant to my weak areas.\\Q6 The
questions were pitched at an appropriate difficulty.\\Q7 The ``why this question'' note made
the targeting understandable.
\item[] \emph{Feedback usefulness.}\\Q8 The feedback clearly identified my mistakes/misconceptions.\\Q9 The feedback helped me understand how to improve.
\item[] \emph{Transparency (co-learning).}\\Q10 Seeing the mastery estimate update after each answer
built my trust.\\Q11 I could see the tutor building a model of what I know.
\item[] \emph{Overall.}\\Q12 Overall, CoLearn was useful for learning.
\item[] \emph{Adoption.}\\Q13 I would use CoLearn again / recommend it.
\end{itemize}
 
\section{Generation \& grading prompts}
\label{sec:appendix-prompts}
The boxes below give the system prompts used by the live system, lightly
reformatted for the single-column width (code identifiers are shown in
\texttt{\{\}}; the \texttt{\{subject\_label\}} slot is filled with the active
subject, e.g.\ ``AQA GCSE Biology'').
 
\begin{promptbox}{Grading: the LLM observation function}
\footnotesize
\emph{Role.}~An expert \texttt{\{subject\_label\}} answer analyst that reads one
student answer at a time and extracts evidence for a topic-level memory profile.
\smallskip
 
\emph{Core rules.}
\begin{itemize}[leftmargin=1.1em,itemsep=2pt,topsep=2pt]
\item Output only valid JSON matching the schema below.
\item \texttt{\{mastery\_evidence\}} and \texttt{\{evidence\_strength\}} are floats in $[0,1]$.
\item \texttt{\{label\_summaries\}} and \texttt{\{label\_evidence\_quotes\}} have equal
length and correspond index-for-index.
\item The predicted label list starts \textsc{empty} and grows only from clear
evidence in the answer.
\end{itemize}
 
\emph{Two label types.}
\begin{itemize}[leftmargin=1.1em,itemsep=2pt,topsep=2pt]
\item \textbf{Positive} --- the student stated something factually wrong; the
evidence quote is a verbatim phrase copied from \texttt{student\_answer}.
\item \textbf{Absence} --- an omission or a too-vague answer; the evidence quote
is either the full short answer or the literal token
\texttt{\textless{}missing: "the specific omitted idea"\textgreater{}}. Never
invent a quote not in the answer.
\end{itemize}
 
\emph{Canonical preference.}~When a \texttt{candidate\_label\_summaries} entry
matches the detected misconception in meaning, copy that candidate string
verbatim; invent new wording only when no candidate fits. \emph{No two labels may
share an evidence quote} (or have one be a substring of another).
\smallskip
 
\textbf{Schema:}~\schema{\{ "topic": ..., "mastery\_evidence": 0.0,
"evidence\_strength": 0.0, "label\_summaries": [...], "label\_evidence\_quotes":
[...], "rationale": ... \}}
\end{promptbox}
 
\begin{promptbox}{Personalised question generation (short-answer)}
\footnotesize
\emph{Role.}~Writes a new \texttt{\{subject\_label\}}-style diagnostic question
targeting a weak topic for a specific student and, when supplied, one or two
target diagnostic error labels.
\smallskip
 
\emph{Rules.}
\begin{itemize}[leftmargin=1.1em,itemsep=2pt,topsep=2pt]
\item Output only valid JSON.
\item Write a concise exam-style question; include a mark scheme or expected
answer.
\item Make it diagnostic: a weak answer should expose the target error or vague
understanding.
\item Use GFM Markdown for tables/lists and \LaTeX{} for equations.
\item Do not mention AI generation; do not include hidden mastery scores.
\end{itemize}
 
\textbf{Schema:}~\schema{\{ "question": ..., "mark\_scheme": ...,
"expected\_diagnostic\_signal": "what a weak/incorrect answer should reveal" \}}
\end{promptbox}
 
\begin{promptbox}{Misconception-tagged MCQ generation}
\footnotesize
\emph{Role.}~Writes a new \texttt{\{subject\_label\}}-style multiple-choice
question from a topic, a student profile, and a list of target misconceptions.
\smallskip
 
\emph{Rules.}
\begin{itemize}[leftmargin=1.1em,itemsep=2pt,topsep=2pt]
\item Output only valid JSON; exactly one correct option.
\item Provide one distractor for \textsc{each} supplied misconception; a student
holding that misconception should be tempted to pick it.
\item Map each distractor to the exact misconception text it tests.
\item Keep options parallel in length/style; do not reveal the correct option or
mention AI.
\end{itemize}
 
\textbf{Schema:}~\schema{\{ "question": ..., "correct\_option": ...,
"distractors": [\{"text": ..., "misconception": ...\}],
"expected\_diagnostic\_signal": "what choosing a distractor reveals" \}}
\end{promptbox}
 
A fourth prompt plans whether two misconceptions can be tested together
(\emph{same mechanism}, \emph{cause--effect}, \emph{commonly-confused contrast},
etc.) before generation; it is omitted here for space.
 
\begin{promptbox}{Diagnostic-quality judge (0--3)}
\footnotesize
\emph{Role.}~An expert \texttt{\{subject\_label\}} question-quality judge that
scores whether a generated question matches its intended weakness target.
\smallskip
 
\emph{Rules.}
\begin{itemize}[leftmargin=1.1em,itemsep=2pt,topsep=2pt]
\item Output only valid JSON.
\item Choose the topic the question actually assesses, and only the supplied
label summaries it could diagnose.
\item Score diagnostic quality: \textbf{0} not diagnostic; \textbf{1} loosely
related but weak; \textbf{2} clearly tests the topic; \textbf{3} clearly exposes
the target diagnostic error label.
\end{itemize}
 
\textbf{Schema:}~\schema{\{ "judged\_topic": ..., "judged\_label\_summaries":
["exact supplied label summary"], "diagnostic\_quality\_score": 0, "rationale":
... \}}
\end{promptbox}
 
\section{Worked co-learning trace}
\label{sec:appendix-trace}
One real round captured from the live deployment (AWS Bedrock, Claude Sonnet 4.5), GCSE Biology,
skill \emph{Levels of organisation within an ecosystem}.
 
\begin{quote}\small
\textbf{Question.} ``Water is an abiotic factor. Name one biotic factor which may affect the
number of buttercups growing on the field.''\\[2pt]
\textbf{Learner answer.} ``Water is a biotic factor, and biotic and abiotic factors basically mean
the same thing.''\\[2pt]
\textbf{Grader} (observation function) $\rightarrow$ \texttt{mastery\_evidence}=0.0; mined
misconceptions: (i) ``incorrectly states water is a biotic factor''; (ii) ``incorrectly states
biotic and abiotic factors mean the same thing''; (iii) ``fails to name any biotic factor''.\\[2pt]
\textbf{Feedback shown.} ``Not quite yet --- let's work on this together.''\\[2pt]
\textbf{Memory update.} mastery $0.56 \rightarrow 0.23$ (the competence-gated BKT lowers the
estimate because the answer is clearly wrong; the misconceptions are stored on the skill).\\[2pt]
\textbf{Next turn.} this skill is now the learner's weakest, so the adaptive engine re-targets it;
on the follow-up item the learner answered correctly and the estimate recovered
($0.23 \rightarrow 0.56$).
\end{quote}
This is the loop in miniature: a graded answer becomes structured evidence, the agent's belief
moves, and the next question is chosen from that updated belief.

\end{document}